\documentclass{article}

\PassOptionsToPackage{numbers,compress}{natbib}
\usepackage[dblblindworkshop, final]{neurips_2026}
 \workshoptitle{Linguistic Principles for Foundation Models (LP4FM)}
\usepackage[utf8]{inputenc} % allow utf-8 input
\usepackage[T1]{fontenc}    % use 8-bit T1 fonts
\usepackage{hyperref}       % hyperlinks
\usepackage{url}            % simple URL typesetting
\usepackage{booktabs}       % professional-quality tables      % 
\usepackage{nicefrac}       % compact symbols for 1/2, etc.

\usepackage{amsmath,amssymb,amsfonts}
\usepackage{algorithmic}
\usepackage{graphicx}
\usepackage{textcomp}
\usepackage{natbib}

\usepackage{placeins}
\usepackage{caption}
\usepackage{subcaption}
\usepackage{wrapfig}
\usepackage{xcolor}

\usepackage{longtable}
\usepackage{eqparbox}
\usepackage{float}
\usepackage{lipsum}
\usepackage{multirow}
\usepackage{scalerel}
\usepackage{mathtools}
\usepackage{microtype}
\usepackage{enumitem}
\usepackage[capitalize]{cleveref}
\Crefname{Chapter}{Chap.}{Chaps.}
\Crefname{Section}{Sec.}{Secs.}
\Crefname{Figure}{Fig.}{Figs.}

\usepackage{comment}

\usepackage{glossaries}
\glsdisablehyper
\newacronym{asr}{ASR}{automatic speech recognition}
\newacronym{cefr}{CEFR}{Common European Framework of Reference for Languages}
\newacronym{ctc}{CTC}{connectionist temporal classification}
\newacronym{gop}{GoP}{Goodness-of-Pronunciation}
\newacronym{hmm}{HMM}{Hidden Markov Model}
\newacronym{ssl}{SSM}{self-supervised speech model}
\newacronym{svd}{SVD}{singular value decomposition}

\newacronym{sandi}{S\&I}{Speak and Improve Corpus 2025}
\newacronym{timit}{TIMIT}{TIMIT}

\def\pn{$\mathcal{P}_{\scaleto{\text{native}}{3.5pt}}$}
\def\pl{$\mathcal{P}^{r}_{\scaleto{\text{L}2}{3.5pt}}$}

\def\cl{$C^{r}_{\scaleto{\text{L}2}{3.5pt}}$}
\def\cnativestar{$C^{\star}_{\scaleto{\text{native}}{3.5pt}}$}
\def\clstar{$C^{r,\star}_{\scaleto{\text{L}2}{3.5pt}}$}
\def\cnativestarp{$C^{\star}_{\scaleto{\text{native},p}{3.5pt}}$}
\def\clstarp{$C^{r,\star}_{\scaleto{\text{L2,p}}{3.5pt}}$}
\def\pc{phone-class}

\makeatother

\title{A Native-Reference Coordinate Geometry for \\ L2 Pronunciation Deviation Using  \\ Self-Supervised Speech Models}

\author{%
	Tina Raissi \quad Nhan Phan \quad Mikko Kurimo \\
	Department of Information and Communications Engineering,
	Aalto University,
	Espoo, Finland \\
	\texttt{\{firstname\}.\{lastname\}@aalto.fi} \\
}

\begin{document}

	\maketitle

		\begin{abstract}
			%We investigate whether pronunciation-related information can be extracted from pretrained self-supervised speech representations without training a pronunciation assessment model or a level predictor. 
			Self-supervised speech models encode rich phonetic information, but it remains unclear how to transform this information into interpretable metrics for second-language~(L2) pronunciation assessment in spontaneous speech.
			We propose a native-reference coordinate geometry in which phone-class averages from native speech define a low-dimensional reference subspace, and L2 speech is evaluated by its distance to matching native phone-class coordinates. 
			Unlike prior distance-based approaches, our method does not require parallel recordings with matched linguistic content or dedicated pronunciation labels.
			%Unlike prior distance-based approaches, our method does not require parallel recordings or matched linguistic content between native and L2 speakers, nor does it require dedicated pronunciation labels.
			Across different self-supervised encoders and modeling choices, the resulting native-reference distances show negative Spearman correlations up to $-0.5$ with speaking proficiency, indicating that higher-proficiency speakers tend to have smaller distances to the corresponding native phone-class coordinates.
			%These results suggest that native-reference geometry provides a label-free pronunciation-related signal and offers a complementary alternative to supervised pronunciation scoring methods trained on human pronunciation ratings.

		%Self-supervised speech models encode rich phonetic information, but it remains unclear how to transform this information into interpretable metrics for second-language pronunciation assessment. We propose a native-referenced distance measure based on self-supervised representations limited to context-dependent phone-classes. Using native speech, we average frame-level self-supervised features for each phone-class using an alignment and construct a low-rank native-reference space with singular value decomposition. For each L2 utterance, we compute corresponding phone-context representations, project them into the native space, and measure their distance to the native coordinates. We test the validity of the distances by asking whether it correlates with the proficiency levels. Across distance metrics, representation layers of different encoders, and different phone-class and feature-slicing methods, we find that higher-proficiency speakers tend to have smaller native-reference distances. These results suggest that self-supervised phonetic geometry can provide an interpretable representation for L2 pronunciation assessment.

	\end{abstract}

\section{Introduction and Motivation}
\label{sec:intro}

Second language~(L2) speaking proficiency is commonly described using the \gls{cefr}, which defines ordered proficiency levels from A1 to C2~\citep{CEFR_2001}. 
Within the \gls{cefr} framework, phonological competence is one component of communicative language ability and is captured by a dedicated phonological control scale. 
Automatic speaking assessment systems have been developed to predict holistic L2 speaking proficiency from learner speech, including scores aligned with \gls{cefr} proficiency levels~\citep{al-ghezi_ASA2023a,qian2025speak}.
However, a holistic proficiency prediction does not necessarily reveal how pronunciation-related information is represented in the underlying speech model. 
The pronunciation assessment based on \Gls{gop} evaluates pronunciation at the phone level by comparing an observed realization with its expected realization under an acoustic model~\citep{el2023automatic}. 
 However, this type of assessment has focused on read-aloud or otherwise constrained speech with known transcriptions~\citep{kyriakopoulos_phone_distrance_2018}, making its broader application to spontaneous speech less straightforward.

In recent years, various studies have investigated the phonetic information captured in self-supervised speech models~\citep{pasad2021,yang21c_interspeech,wells2022phonetic,pasad2023comparative}.
Further evidence comes from analyses of representations extracted from phone-aligned segments, which show that the self-supervised features encode context-dependent phone information~\citep{choi2025leveraging,choi2026b}.
 In particular, these representations are not restricted to information from the corresponding phone, but also incorporate larger context~\citep{pasad2024self}.
However, these findings do not directly yield an interpretable metric for L2 pronunciation assessment.

Prior work has investigated distance measures based on dynamic time warping alignments between L2 and native speech, represented by conventional Mel-cepstral or self-supervised features~\citep{bartelds2020new,chernyak2024perceptual,mcintosh2026self}. However, these approaches require parallel recordings of the same text from both speakers and may be sensitive to disfluencies, which are common in L2 speech.
We propose an approach that does not require matched linguistic content.
%enabling distance based comparison between L2 and native speech from non-parallel utterances.
To avoid the need for utterance-level parallelism we use a \pc{} level comparison, where L2 and native speech are compared only through shared classes, after projecting their representations into a common native-derived subspace.
For this purpose, we investigate whether there exists a low-dimensional geometric subspace within self-supervised feature spaces where, (1) L2 phone realizations can be anchored to native-reference phone representations, and (2) a distance measure can quantify L2 pronunciation deviation.
Using self-supervised representations of utterances of a native corpus we accumulate \pc{} averaged representations using segment boundary information via the Viterbi alignment from an external frame-wise model.
We then introduce a low-rank native \pc{} coordinate space and its corresponding projection basis by performing \gls{svd}.
 For each L2 utterance, we project its \pc{} averages into the native projection basis, yielding L2 \pc{} coordinates.
Pronunciation deviation is measured as the distance between L2 and native-reference coordinates for matching \pc es, using different distance measures.
Notably, we do not train a model using the representations or the distances to predict the proficiency level.
Instead, we test the raw distance validity of the proposed method.
We propose that if the native-reference subspace captures pronunciation-relevant information, then speakers with higher proficiency level should, on average, have lower native-reference distances.
We evaluate this hypothesis using Spearman's rank correlation coefficient~\citep{spearman_1904}.
Moreover, we analyze how the correlation between the distance and proficiency level is affected by the choice of \pc{}, averaging method, and subspace rank dimension, as well as the choice of self-supervised model and the encoder layer.

\begin{table}[t]
	\centering
	\small
	\setlength{\tabcolsep}{0.9em}
	\renewcommand{\arraystretch}{1.1}
		\caption{Pretrained self-supervised speech encoders used in this work.
		We report the number of parameters~(\#PMs), the language and number of hours~(\#H) for the training data, and probed layers.}
			\label{tab:ssl_models}
	\begin{tabular}{| l | c | c | c | c|}
		\hline 
	\multirow{2}{*}{\textbf{Model}}
		& \multirow{2}{*}{\textbf{\#PMs}}
		& \multicolumn{2}{c|}{\textbf{Training Data}}
		& \multirow{2}{*}{\textbf{Probing Layers}} \\ 
		\cline{3-4}
		&
		&
		\textbf{\#H}
		& \textbf{Language}
		& \\
		\hline
		
\bf	WavLM Base+
		& $\sim$95M
		& \multirow{2}{*}{94k}
		& \multirow{3}{*}{English}
		& \footnotesize \{6, 9\}
		\\
		\cline{1-2} \cline{5-5}
		
	\bf 	WavLM Large
		& \multirow{3}{*}{$\sim$317M}
		&
		&
		& \{6, 9, 12, 15\}
		\\
		\cline{1-1} \cline{3-3} \cline{5-5}
		
	\bf	wav2vec 2.0 Large (LV-60)
		&
		& 60k
		&
		&\multirow{2}{*}{ \{12, 15\}}
		\\
		\cline{1-1} \cline{3-4}
		
	\bf	wav2vec 2.0 XLSR-53
		&
		& 56k
		& Multilingual
		&
		\\ \hline
		
	\end{tabular}
\vspace{-0.3cm}
\end{table}
\begin{table}[t]
	\centering
	\small
	\setlength{\tabcolsep}{0.3em}
	\renewcommand{\arraystretch}{1.0}
		\caption{Different \pc es with center-phone averaging strategy on the output of the middle layer of each encoder, using \gls{svd} rank of 16.}
	\label{tab:phon_duration_correlation}
	\begin{tabular}{|l|ccc|ccc|ccc|}
		\hline
		\multirow{2}{*}{\textbf{Encoder}}
		& \multicolumn{3}{c|}{\textbf{Monophone}}
		& \multicolumn{3}{c|}{\textbf{Diphone}}
		& \multicolumn{3}{c|}{\textbf{Triphone}} \\
		\cline{2-10}
		& \textbf{Euc}
		& \textbf{Maha}
		& \textbf{Cos}
		& \textbf{Euc}
		& \textbf{Maha}
		& \textbf{Cos}
		& \textbf{Euc}
		& \textbf{Maha}
		& \textbf{Cos}
		\\
		\hline \hline
		
		\textbf{WavLM Base+}
		& \textbf{-0.3893} & -0.3861 & -0.3183
		& \textbf{-0.4337} & -0.4329 & -0.3888
		& -0.4691 & \textbf{-0.4707} & -0.4467
		\\
		
		\hline
		
		\textbf{WavLM Large}
		& -0.3155 & \textbf{-0.3197} & -0.2939
		& -0.3956 & \textbf{-0.3984} & -0.3759
		& -0.4616 & \textbf{-0.4657} & -0.4448
		\\
		
		\hline
		
		\textbf{wav2vec 2.0 Large}
		& -0.1358 & -0.0990 & \textbf{-0.2730}
		& -0.1852 & -0.1823 & \textbf{-0.3748}
		& -0.2302 & -0.2578 & \textbf{-0.4001}
		\\
		
		\hline
		
		\textbf{wav2vec 2.0 XLSR-53}
		& -0.1798 & -0.1378 & \textbf{-0.2431}
		& -0.2272 & -0.2605 & \textbf{-0.3166}
		& -0.2855 & -0.3344 & \textbf{-0.3689}
		\\
		
		\hline
	\end{tabular}
\vspace{-0.3cm}
\end{table}

\begin{table}[t]
	\centering
	\small
	\setlength{\tabcolsep}{1.1em}
	\renewcommand{\arraystretch}{1.0}
		\caption{Similar experiments as in \cref{tab:phon_duration_correlation}, showing the effect of averaging strategies for triphone.}
		\label{tab:triphone_strategy}
	\begin{tabular}{|l|ccc|ccc|}
		\hline
		\multirow{2}{*}{\textbf{Encoder}}
		& \multicolumn{3}{c|}{\textbf{Center-Phone}}
		& \multicolumn{3}{c|}{\textbf{Full-Unit}} \\
		\cline{2-7}
		& \textbf{Euc}
		& \textbf{Maha}
		& \textbf{Cos}
		& \textbf{Euc}
		& \textbf{Maha}
		& \textbf{Cos}
		\\
		\hline \hline
		
		\textbf{WavLM Base+}
		& -0.4691
		& -0.4707
		& -0.4467
		& -0.4417
		& -0.4432
		& \bf{-0.4995}
		\\
		\hline
		
		\textbf{WavLM Large}
		& -0.4616
		& -0.4657
		& -0.4448
		& -0.4804
		& -0.4725
		& \bf{-0.4970}
		\\
		\hline
		
		\bf{wav2vec 2.0 Large}
		& -0.2302
		& -0.2578
		& \bf{-0.4001}
		& -0.1817
		& -0.1907
		& -0.3569
		\\
		\hline
		
		\textbf{wav2vec 2.0 XLSR-53}
		& -0.2855
		& -0.3344
		& \bf{-0.3689}
		& -0.2191
		& -0.2440
		& -0.3168
		\\
		\hline
		
	\end{tabular}
\vspace{-0.2cm}
\end{table}

\begin{table}[t]
	\centering
	\small
	\setlength{\tabcolsep}{0.45em}
	\renewcommand{\arraystretch}{1.}
		\caption{Spearman coefficients using Cosine distance for different \gls{svd} ranks. All experiments use triphone with full-unit averaging using the representations of the encoder's middle layer.}
	\label{tab:triphone_rank_correlation}
	\begin{tabular}{|l|cc|cc|cc|cc|cc|}
		\hline
		\multirow{2}{*}{\textbf{WavLM}}
		& \multicolumn{2}{c|}{\textbf{Rank 8}}
		& \multicolumn{2}{c|}{\textbf{Rank 16}}
		& \multicolumn{2}{c|}{\textbf{Rank 32}}
		& \multicolumn{2}{c|}{\textbf{Rank 64}}
		& \multicolumn{2}{c|}{\textbf{Rank 128}} \\
		\cline{2-11}
		& \textbf{Train} & \textbf{Dev}
		& \textbf{Train} & \textbf{Dev}
		& \textbf{Train} & \textbf{Dev}
		& \textbf{Train} & \textbf{Dev}
		& \textbf{Train} & \textbf{Dev} \\
		\hline
		
		\textbf{Base+}
		& -0.4711 & -0.4502
		& -0.4995 & -0.4723
		& -0.5092 & -0.4898
		& \textbf{-0.5151} & \textbf{-0.4948}
		& -0.5107 & -0.4878 \\
		\hline \hline
		
		\textbf{Large}
		& -0.4945 & -0.4765
		& -0.4970 & -0.4701
		& -0.4958 & -0.4725
		& \textbf{-0.4974} & \textbf{-0.4773}
		& -0.4957 & -0.4727 \\
		\hline
		
	\end{tabular}
\vspace{-0.2cm}
\end{table}

\begin{table}[t]
	\centering
	\small
	\setlength{\tabcolsep}{0.4em}
	\setlength{\abovecaptionskip}{5pt}
	\renewcommand{\arraystretch}{1.}
		\caption{Effect of factored hybrid~(FH) HMM and CTC alignment models for full-unit triphone using features from middle layer of WavLM Base+, when using \gls{svd} rank of 64 and cosine distance.}
	\begin{tabular}{|l|cc|cc|cc|cc|cc|}
		\hline
		\textbf{Alignment}
		& \multicolumn{2}{c|}{\textbf{Part 1}}
		& \multicolumn{2}{c|}{\textbf{Part 3}}
		& \multicolumn{2}{c|}{\textbf{Part 4}}
		& \multicolumn{2}{c|}{\textbf{Part 5}}
		& \multicolumn{2}{c|}{\textbf{Fisher Average}} \\
		\cline{2-11}
		& \textbf{Train} & \textbf{Dev}
		& \textbf{Train} & \textbf{Dev}
		& \textbf{Train} & \textbf{Dev}
		& \textbf{Train} & \textbf{Dev}
		& \textbf{Train} & \textbf{Dev} \\
		\hline
		
		\textbf{FH HMM}
		& \textbf{-0.5399} & \bf -0.4000
		& \textbf{-0.5399} & \textbf{-0.5647}
		& -0.5012 & \textbf{-0.4969}
		& \textbf{-0.4775} & \textbf{-0.5087}
		& \textbf{-0.5151} & \textbf{-0.4948} \\
		\hline 
		
		\textbf{CTC}
		& -0.4612 & -0.3681
		& -0.5261 & -0.5249
		& \textbf{-0.5042} & -0.4891
		& -0.4126 & -0.4757
		& -0.4772 & -0.4663 \\
		\hline
		
	\end{tabular}
\vspace{-0.4cm}
	\label{tab:alignment_correlation}
\end{table}

\section{Method Description}
\label{sec:method}
 Let $H^{\ell} \in \mathbb{R}^{T \times D}$ denote the output feature sequence of the $\ell$-th layer of the encoder in a self-supervised model. 
 Define $\Phi$ of length $N$ as the output phonetic label sequence representing the spoken word sequence. 
 Each element in $\Phi$ is an ARPAbet phone~\citep{shoup1980phonological}.
 Let $\mathcal{P}$ be the finite set of \emph{\pc es},
 and define a phonetic alignment sequence $\mathcal{A} = \{a_{1},\dots,a_{T}\}$.
At time frame $t$, $a_t$ aligns a \pc{} $p\in\mathcal{P}$, such as monophone, diphone, or triphone to the encoder output $h_t$.

\subsection{Building Native-Reference and L2 Coordinates}
\label{subsec:codebook}
Define $\mu(p)=\frac{1}{|\mathcal{T}_p|}\sum_{t\in\mathcal{T}_p} h_t$ the {\pc{} mean vector} for class $p$, where $\mathcal{T}_p \coloneqq \{t: a_t = p\}$.
We build a native matrix $M\in\mathbb{R}^{\vert\text{\pn}\vert\times D}$ where \pn correspond to the set of \pc es derived from the transcriptions of the native speech corpus.
Each row of the native matrix \(M_p=\mu(p)^{\top}\) represents the mean vector \(\mu(p)\) associated with a single \pc{} \(p\in\text{\pn}\).
%Each row of the native matrix $M_{p}=\mu(p)^{\top}$ correspond to one \pc{} mean vector  $p\in\text{\pn}$.
We introduce a global native mean vector $\bar{\mu}_{\text{native}}=\tfrac{1}{\vert\mathcal{\text{\pn}}\vert}\sum_{p\in\text{\pn}}\mu(p)$, and denote $\bar{M}= M - \mathbf{1} \bar{\mu}^{\top}_{\text{native}}$ as the centered \pc{} mean matrix.

We perform rank-$K$ truncated \gls{svd} of the centered native \pc{} mean matrix
\begin{equation}\nonumber \footnotesize
	\bar{M}\approx U\Sigma V^{\top},\quad
	U\in\mathbb{R}^{\vert\text{\pn}\vert\times K},\;\Sigma\in\mathbb{R}^{K\times K},\;V\in\mathbb{R}^{D\times K},
\end{equation}
where $U\Sigma V^{\top}$ is the best possible rank-$K$ linear approximation of the centered native \pc{} mean matrix under Frobenius norm.

The \emph{projection basis matrix} $V$ contains orthonormal directions spanning a $K$-dimensional native-reference subspace of the original self-supervised feature space, capturing dominant variation among centered native \pc{} mean vectors.

Define $C_{\text{native}}=U\Sigma\in\mathbb{R}^{\vert\mathcal{\text{\pn}}\vert\times K}$ to be our \emph{native-reference coordinate}.
This is a low-dimensional coordinate space of \pc{} vectors in a native-reference subspace.

For an L2 utterance $r$ and its corresponding self-supervised feature sequence $H_r^{\ell} \in \mathbb{R}^{T_r \times D}$, we build $M^{r}\in\mathbb{R}^{\vert\mathcal{\text{\pl}}\vert \times D}$ in the same way as for $M$ with the difference that now \pl{} is the set \pc es appearing in the transcript of the L2 utterance $r$.
The corresponding \emph{L2 coordinate} is then calculated by projecting $M^{r}$ into the native \gls{svd} subspace $V$, after subtracting the native global mean $\bar{\mu}_{\text{native}}$, as shown in \cref{eq:l2-coordinate}.\vspace{-0.3cm}
\vspace{-0.05cm}
\begin{equation}\label{eq:l2-coordinate} \footnotesize
\text{\cl} = (M^{r} -  \mathbf{1} \bar{\mu}^{\top}_{\text{native}} ) V
\end{equation}

\subsection{Native-reference distance definition}
\label{subsec:distance}
For each utterance $r$, we restrict the native and L2 coordinate matrices to the common \pc es
$\mathcal{P}^{\star}=\text{\pn}\cap\text{\pl}$, yielding \cnativestar and \clstar.
For each $p \in \mathcal{P}^{\star}$, we compute a \pc{} level distance
$
\delta_r(p) =
d_m(
\text{\cnativestarp},
\text{\clstarp}
)
$, for distance measure $d_m$.
The utterance-level native-reference distance is then obtained by averaging over the common \pc es.
%For each utterance $r$, we first select the set of common \pc es $\mathcal{P}^{\star}=\ \text{\pn} \cap  \text{\pl}$.
%We denote by \cnativestar and \clstar a native and an L2 coordinate matrix restricted to common \pc es.
%For each $p \in \mathcal{P}^{\star}$, a \pc{} distance is computed by $\Delta_{r}= d_{\mathrm{m}}(p)(\text{\cnativestar}, \text{\clstar})$, where $m$ denotes the chosen distance measure.
%An overview of the distance measures used in this work is reported in \cref{tab:distances}.
We selected three distance measures to capture complementary aspects of L2-native phone-coordinates deviations: (1) Euclidean distance measures absolute displacement in the projected \gls{svd} subspace, (2) cosine distance captures angular differences independent of magnitude, and (3) a regularized diagonal Mahalanobis variant scales coordinate differences using variances estimated across native \pc{} coordinates (see \cref{tab:distances}).

\begin{figure}[h!]
	\centering
	\includegraphics[height=0.65\linewidth]{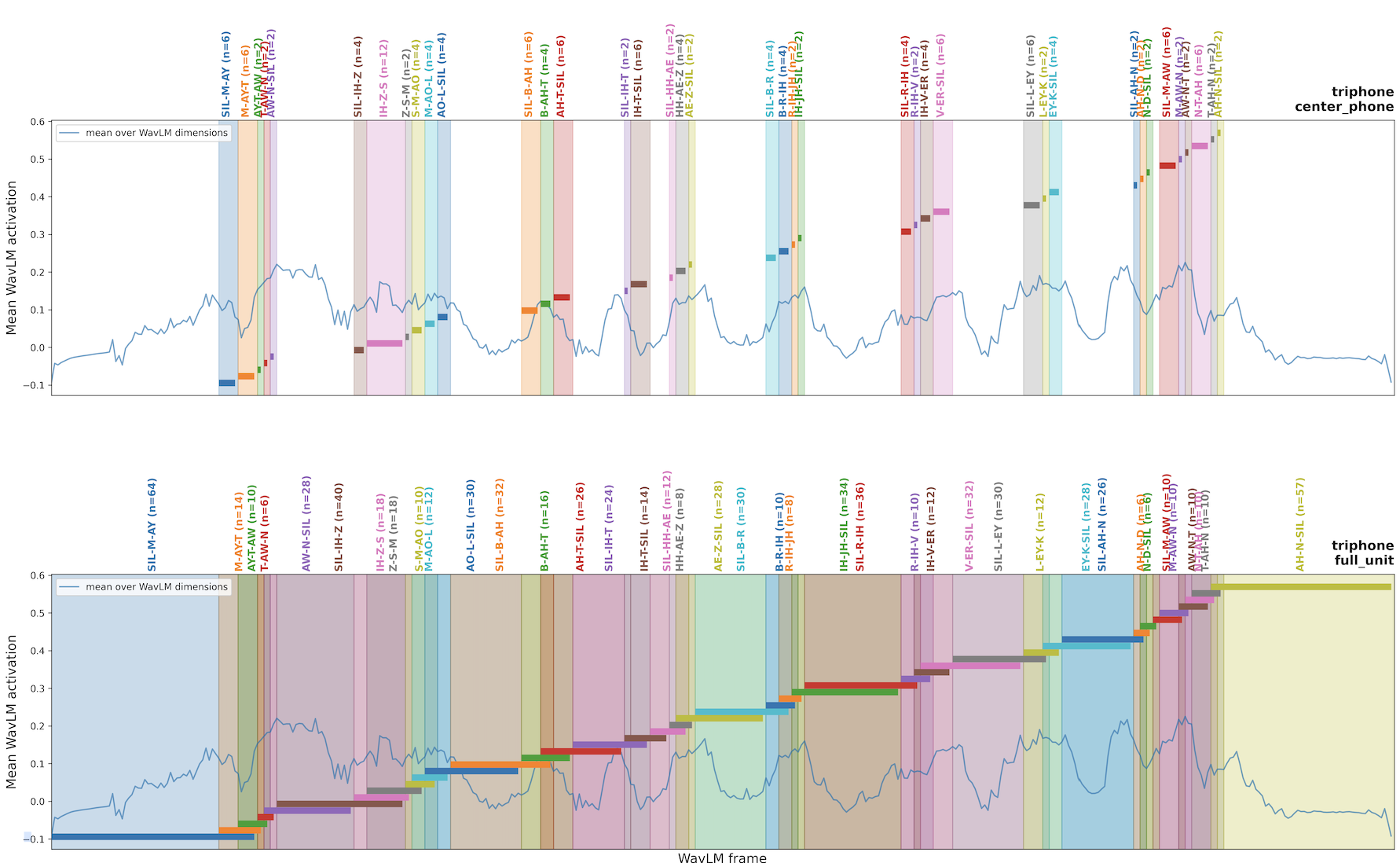}
	\caption{Comparison of center-phone and full-unit averaging strategies for a triphone phone-class.
	We show the aligned triphone boundaries and segments selected for averaging.
	For simplicity, we show the averaged value over feature dimension for the WavLM output in the blue plot.
	The utterance is selected from train data of \gls{sandi}.}
	\label{fig:triphonecomparison}
\end{figure}

\paragraph{Averaging Strategies and Phone-Classes:} We use feature-slicing~\cite{pasad2021} over \pc{} segment boundaries provided by an alignment.
For the common monophone, diphone and triphone \pc es, we adopt two different averaging strategies, based on how the segment boundaries are defined.
While \emph{center-phone} averaging considers feature vectors aligned to the center phone of each unit, \emph{full-unit} averaging aggregates in addition the frames aligned to all phones in the \pc{} unit.
A comparison of the two averaging strategies for a triphone \pc{} is depicted in \cref{fig:triphonecomparison} of \cref{sec:app}.
Note that the two strategies coincide for monophones, since center-phone and full-unit pooling both reduce to averaging the frames aligned to the single phone.

\paragraph{Evaluation Metric}
We evaluate the association between native-reference distances $\Delta$ and the \gls{cefr}-aligned speaking proficiency using Spearman's rank correlation coefficient $\rho$~\citep{spearman_1904}.
We verify whether higher-proficiency L2 speakers tend to have lower native-reference distances by measuring the monotonic association between the distance and CEFR based on their relative ordering. 
Lower distances indicate greater similarity to the native-reference coordinates, while higher \gls{cefr} scores indicate greater proficiency. 
Therefore, a negative Spearman correlation would indicate that native-reference distance decreases as proficiency increases, supporting an inverse association between the distance and the \gls{cefr}.
As a rule of thumb for rank correlations with human assessment data, $|\rho|\approx 0.3$ is often treated as a moderate association, while $|\rho|\approx 0.5$ indicates a meaningful monotonic relationship, especially when the method is not directly trained on the target ratings. 

\section{Experiments}
\label{sec:experiments}

\subsection{Models and Setting}
%For our experiments we use the following models.
%\paragraph{Self-Supervised Models for Probing}
%We use four pretrained self-supervised models belonging to WavLM \todo{cite} and wav2vec2.0 families. 
%The encoder architecture for all models rely on a convolutional front-end followed by a Transformer\todo{cite} block.
%\todo{training criterion, regularizations}
%We used feature sequences extracted from different layers of each pretrained model.
We use TIMIT~\cite{garofolo1993timit} as our native speech corpus together with its official phone segment boundaries.
For L2 speech we use train and dev partitions of the \gls{sandi} corpus~\cite{knill25_slate,qian2025speak}.
For the train data, only utterances with manual transcriptions are used.
While modeling aspects such as averaging strategies and choices of \pc{} are evaluated using the train data, all other experiments additionally use the dev data.
Each subset is divided into four different proficiency evaluation tasks.
For reporting the evaluation results, we first calculate Spearman's coefficient for each part and then average the values, following common practice in the literature.
Here, we report the Fisher average~\cite{fisher1915frequency}.

We extract self-supervised features from different layers of pretrained models belonging to WavLM~\citep{chen2022wavlm} and wav2vec2.0~\citep{baevski2020wav2vec} families. 
A summary of the model parameters and the used layers is reported in \cref{tab:ssl_models}.
%We use the arpabet phones and merge stressed phones into a unique class for a total of 39 phones and one silence label.
%\paragraph{Supervised Models for Alignment}
We use two neural based frame-wise alignment models.
% trained with \gls{hmm} and \gls{ctc} label topologies using phone units.
As the primary model, we used a Conformer based factored hybrid \gls{hmm}~\cite{raissi24_interspeech}, trained on the small subset of the Loquacious~\cite{parcollet25_interspeech,rossenbach2026supplementary},
and a wav2vec2.0 \gls{ctc} fine-tuned on TIMIT \footnote{\tiny \url{https://huggingface.co/speech31/wav2vec2-large-english-TIMIT-phoneme_v3}}.

 %using both frame-wise cross-entropy and fine-tuned by summation over all alignment paths~(full-sum) \todo{cite}. In addition, we also used a wav2vec2.0 Large fine-tuned on TIMIT with full-sum criterion using \gls{ctc} label topology \todo{give reference to model on HF}.
%All models except for factored hybrid HMM are publicly available under HuggingFace.
We used Lhotse~\cite{zelasko2021lhotse} for data preparation and RASR~\cite{rybach2011rasr} for forced alignment. 
We will provide the source code used to conduct all experiments\footnote{\tiny \url{https://github.com/aalto-speech/native-L2-coordinate-geometry}}.

\subsection{Results}
\label{sec:results}

In \cref{tab:phon_duration_correlation,tab:triphone_strategy,tab:triphone_rank_correlation,tab:alignment_correlation}, we investigate different modeling and experimental choices and report the Fisher averaged Spearman coefficient on the \gls{sandi} task.
To compare \pc{} and averaging choices, we first fix several conditions: since the rank is mathematically bounded above by the number of \pc es minus one, we set it to 16, which is appropriate for smaller monophone class set. 
We also use center-phone averaging and select the middle encoder layer, consistent with the WavLM layer-wise results in \cref{tab:triphone_layer_correlation} and prior findings~\citep{pasad2021}.
Under these conditions, \cref{tab:phon_duration_correlation} shows that triphone classes perform best, and that the optimal correlation depends on the encoder type. 
We then fix the triphone class and compare averaging strategies in \cref{tab:triphone_strategy}.
Full-unit averaging improves cosine correlations for both WavLM encoders, while its effect on other distances depends on the encoder.
%observing stronger negative Spearman correlations for the full-unit strategy with WavLM encoders and for the center-phone strategy with wav2vec 2.0. 
For rank selection and alignment analysis, we fix the triphone class with full-unit averaging. 
As shown in \cref{tab:triphone_rank_correlation}, the best rank is 64. 
Finally, \cref{tab:alignment_correlation} shows that \gls{hmm}-based alignment leads to stronger negative correlations than \gls{ctc}-based alignment.
% likely because the blank label in the \gls{ctc} topology serves both label repetition and silence, which can lead to less accurate segmentation.
%likely because the ambiguous role of the blank label in the \gls{ctc} topology, as both label repetition and silence, leads to less accurate segmentation. 
Nevertheless, \gls{ctc} alignment still yields reasonable correlations.

There is no directly comparable prior result for our setting, since related studies differ in the assessment target, speech setting, or supervision used.
%For context, listener-rated accentedness and comprehensibility correlate only modestly with overall oral proficiency ($\rho=-0.26$ to $-0.35$)~\citep{tergujeff_2021}, although these are not direct measures of pronunciation. 
For context, listener-rated accentedness and comprehensibility correlate modestly with overall oral proficiency ($\rho=-0.260$ and $-0.345$, respectively)~\citep{tergujeff_2021}.
Conversely, supervised systems on the same \gls{sandi} proficiency task reach Spearman correlations around $0.83$~\citep{qian2025speak}, but are trained on proficiency labels and can exploit information beyond pronunciation. Against these contextual rather than direct comparisons, our $\rho \approx-0.5$ with a 95\% bootstrap confidence interval of $[-0.5294, -0.4591]$ indicates a clear negative association: the observed correlations suggest that native-reference distances capture information associated with holistic proficiency, without training a supervised proficiency predictor.

\section{Conclusions and Future Work}
We proposed a native-reference coordinate geometry for analyzing L2 pronunciation deviation in self-supervised speech representations. 
Unlike previous approaches, our approach does not require matched text between L2 and native speech.
We construct low-rank coordinate matrices from native and L2 \pc{} averages of self-supervised features, and show that the resulting distances yield negative Spearman correlations of up to around $-0.5$ with speaker proficiency level, suggesting that our proposed method captures meaningful proficiency-related information.
Our results further show that the choice of \pc{}, averaging strategy, rank, encoder, and alignment affects the strength of this correlation. 
Future work will further investigate the methodology, robustness across additional corpora, and the effect of alignment.
	 
	\begin{ack}
	 	This work was supported by the Strategic Research Council grant number 355587, 365233, and 373228.
	 	The authors also acknowledge Aalto University’s Science-IT for providing computational resources.
	\end{ack}

    \bibliographystyle{unsrtnat}
    \bibliography{references}
    
    %%%%%%%%%%%%%%%%%%%%%%%%%%%%%%%%%%%%%%%%%%%%%%%%%%%%%%%%%%%%
    \newpage
    \appendix
    
    \section{Supplementary Material}
    \label{sec:app}

    \subsection{Averaging Strategies}
    
    \subsection{Distance Calculation Formula}
            For the Mahalanobis variant, $v_j$ is the variance
    of coordinate $j$ across all native \pc{} averages, shared across classes.
    We limit the influence of small variances by applying a regularization $\tilde{v}_j=\max\{0.9v_j+0.1\tau,10^{-3}\tau\}$,
    where $\tau$ is the mean positive coordinate variance, and use $\gamma=0.5$. 
    The distance gives greater weight to differences along coordinates
    with smaller native between-class variance.
    
        \begin{table}[h!]
	\centering
	\small
	\setlength{\tabcolsep}{1.1em}
	\renewcommand{\arraystretch}{2.5}
		\caption{Per-class distances between matched L2 and native coordinates
			$x_{rp}=C^r_{\mathrm{L2},p}\in\mathbb{R}^K$ and
			$y_p=C_{\mathrm{native},p}\in\mathbb{R}^K$.
			The utterance score is the mean over shared classes,
			$\Delta_r=|\mathcal{P}^{\star}|^{-1}\sum_{p\in\mathcal{P}^{\star}}d_{\mathrm{m},r}(p)$.}
	\vspace{0.2cm}
	\label{tab:distances}
	
	\begin{tabular}{|l|c|}
		\hline
		\textbf{Distance} & \textbf{Definition} \\
		\hline \hline
		
		Euclidean &
		$\displaystyle
			d_{\mathrm{Euc},r}(p)
			= \sqrt{\frac{1}{K}
				\left\|x_{rp}-y_p\right\|_2^2}
		$ \\
		\hline
		
			Diagonal Mahalanobis variant &
		$\displaystyle
			d_{\mathrm{Maha},r}(p)
			= \sqrt{\frac{1}{K}
				\sum_{j=1}^{K}\frac{(x_{rp,j}-y_{p,j})^2}{\tilde{v}_j^\gamma}}
		$ \\
		\hline
		
		Cosine &
		$\displaystyle
			d_{\mathrm{Cos},r}(p)
				= 1-
			\frac{x_{rp}^{\top}y_p}
				{\|x_{rp}\|_2\|y_p\|_2}
		$ \\
		\hline
	\end{tabular}

\end{table}

        \subsection{WavLM Layer Choice}
        \begin{table}[htbp]
	\centering
	\small
	\setlength{\tabcolsep}{0.9em}
	\setlength{\abovecaptionskip}{5pt}
	\renewcommand{\arraystretch}{1.}
		\caption{Fisher averaged Spearman correlations for different WavLM layers,
			using cosine distance, triphone classes, full-unit averaging, and
			\gls{svd} rank 64. Bold values indicate the most negative correlation
			for each encoder and split.}
		\label{tab:triphone_layer_correlation}
	\begin{tabular}{|l|cc|cc|cc|cc|}
		\hline
		\multirow{2}{*}{\textbf{WavLM}}
		& \multicolumn{2}{c|}{\textbf{Layer 6}}
		& \multicolumn{2}{c|}{\textbf{Layer 9}}
		& \multicolumn{2}{c|}{\textbf{Layer 12}}
		& \multicolumn{2}{c|}{\textbf{Layer 15}} \\
		\cline{2-9}
		& \textbf{Train} & \textbf{Dev}
		& \textbf{Train} & \textbf{Dev}
		& \textbf{Train} & \textbf{Dev}
		& \textbf{Train} & \textbf{Dev} \\
		\hline
		
		\textbf{Base+}
			& \textbf{-0.5151} & \textbf{-0.4948}
		& -0.4848 & -0.4860
		& \multicolumn{4}{c|}{-}
 \\
		\hline \hline
		
		\textbf{Large}
		& -0.4718 & -0.4182
		& -0.4938 & -0.4696
			& \textbf{-0.4974} & \textbf{-0.4773}
		& -0.4672 & -0.4610 \\
		\hline
		
	\end{tabular}

\end{table}

        We observed that for WavLM Base+, layer 6 yields stronger negative
        correlations than layer 9. Similarly, for WavLM Large, layer 12 gives the most
        negative correlations among the tested layers. Both conclusions hold
        on train and dev, supporting the choice of an intermediate layer for each encoder.

	%%%%%%%%%%%%%%%%%%%%%%%%%%%%%%%%%%%%%%%%%%%%%%%%%%%%%%%%%%%%
	
	\newpage
	\section*{NeurIPS Paper Checklist}
 %You should answer \answerYes{}, \answerNo{}, or \answerNA{}.

\begin{enumerate}

\item {\bf Claims}
    \item[] Question: Do the main claims made in the abstract and introduction accurately reflect the paper's contributions and scope?
    \item[] Answer:  \answerYes{} % Replace by \answerYes{}, \answerNo{}, or \answerNA{}.
    \item[] Justification: We propose a method and describe it mathematically and provide the implementation.

\item {\bf Limitations}
    \item[] Question: Does the paper discuss the limitations of the work performed by the authors?
    \item[] Answer:  \answerYes{}% Replace by \answerYes{}, \answerNo{}, or \answerNA{}.
    \item[] Justification: The current study is limited to English and is affected by the domain mismatch between TIMIT and spontaneous S\&I speech, the reliance on phone alignment, and the use of holistic proficiency rather than direct pronunciation ratings.

\item {\bf Theory assumptions and proofs}
    \item[] Question: For each theoretical result, does the paper provide the full set of assumptions and a complete (and correct) proof?
    \item[] Answer:  \answerNA{}% Replace by \answerYes{}, \answerNo{}, or \answerNA{}.
    \item[] Justification: Our paper is based on a mathematical concept but it is not purely theoretical.

    \item {\bf Experimental result reproducibility}
    \item[] Question: Does the paper fully disclose all the information needed to reproduce the main experimental results of the paper to the extent that it affects the main claims and/or conclusions of the paper (regardless of whether the code and data are provided or not)?
    \item[] Answer:  \answerYes{}% Replace by \answerYes{}, \answerNo{}, or \answerNA{}.
    \item[] Justification: The datasets, model configurations, alignment methods, projection settings, and evaluation procedure are described in the paper. We will release the source code and configurations required to reproduce the reported experiments.

\item {\bf Open access to data and code}
    \item[] Question: Does the paper provide open access to the data and code, with sufficient instructions to faithfully reproduce the main experimental results, as described in supplemental material?
    \item[] Answer:  \answerNo{} % Replace by \answerYes{}, \answerNo{}, or \answerNA{}.
    \item[] Justification: TIMIT and S\&I are available for research use under their respective access conditions. The experimental code will be released upon acceptance.

\item {\bf Experimental setting/details}
    \item[] Question: Does the paper specify all the training and test details (e.g., data splits, hyperparameters, how they were chosen, type of optimizer) necessary to understand the results?
    \item[] Answer:  \answerYes{} % Replace by \answerYes{}, \answerNo{}, or \answerNA{}.
    \item[] Justification: These are described in our Experimental section.

\item {\bf Experiment statistical significance}
    \item[] Question: Does the paper report error bars suitably and correctly defined or other appropriate information about the statistical significance of the experiments?
    \item[] Answer:  \answerYes{}% Replace by \answerYes{}, \answerNo{}, or \answerNA{}.
    \item[] Justification: We define our evaluation metric and its reasonable range and provide confidence interval for our best result on dev data.

\item {\bf Experiments compute resources}
    \item[] Question: For each experiment, does the paper provide sufficient information on the computer resources (type of compute workers, memory, time of execution) needed to reproduce the experiments?
    \item[] Answer: \answerNo{} % Replace by \answerYes{}, \answerNo{}, or \answerNA{}.
    \item[] Justification: No models are trained in this work. The experiments consist primarily of inference with pretrained speech encoders, forced alignment, and low-dimensional linear-algebra operations, and therefore have modest computational requirements.
    
\item {\bf Code of ethics}
    \item[] Question: Does the research conducted in the paper conform, in every respect, with the NeurIPS Code of Ethics \url{https://neurips.cc/public/EthicsGuidelines}?
    \item[] Answer:  \answerYes{}% Replace by \answerYes{}, \answerNo{}, or \answerNA{}.
    \item[] Justification: We follow the guidelines.

\item {\bf Broader impacts}
    \item[] Question: Does the paper discuss both potential positive societal impacts and negative societal impacts of the work performed?
    \item[] Answer:  \answerNA{} % Replace by \answerYes{}, \answerNo{}, or \answerNA{}.
    \item[] Justification: The paper is analyzing the capabilities of model based representations for distinguishing L2 from native speakers. There is no harmful societal impact from our work. However, there is no straightforward way to discuss the positive impact in a short paper either. 
    
\item {\bf Safeguards}
    \item[] Question: Does the paper describe safeguards that have been put in place for responsible release of data or models that have a high risk for misuse (e.g., pre-trained language models, image generators, or scraped datasets)?
    \item[] Answer: \answerNA{} % Replace by \answerYes{}, \answerNo{}, or \answerNA{}.
    \item[] Justification:  We do not introduce or release a new high-risk model or dataset requiring additional safeguards.

\item {\bf Licenses for existing assets}
    \item[] Question: Are the creators or original owners of assets (e.g., code, data, models), used in the paper, properly credited and are the license and terms of use explicitly mentioned and properly respected?
    \item[] Answer:  \answerYes{} % Replace by \answerYes{}, \answerNo{}, or \answerNA{}.
    \item[] Justification: All third-party datasets, pretrained models, and software used in this work are properly cited and are used in accordance with their respective licenses and terms of use.

\item {\bf New assets}
    \item[] Question: Are new assets introduced in the paper well documented and is the documentation provided alongside the assets?
    \item[] Answer:  \answerNA{} % Replace by \answerYes{}, \answerNo{}, or \answerNA{}.
    \item[] Justification: We do not release any new asset.

\item {\bf Crowdsourcing and research with human subjects}
    \item[] Question: For crowdsourcing experiments and research with human subjects, does the paper include the full text of instructions given to participants and screenshots, if applicable, as well as details about compensation (if any)? 
    \item[] Answer: \answerNA{} % Replace by \answerYes{}, \answerNo{}, or \answerNA{}.
    \item[] Justification: paper does not involve crowdsourcing.

\item {\bf Institutional review board (IRB) approvals or equivalent for research with human subjects}
    \item[] Question: Does the paper describe potential risks incurred by study participants, whether such risks were disclosed to the subjects, and whether Institutional Review Board (IRB) approvals (or an equivalent approval/review based on the requirements of your country or institution) were obtained?
    \item[] Answer: \answerNA{}  % Replace by \answerYes{}, \answerNo{}, or \answerNA{}.
    \item[] Justification: Not involved.

\item {\bf Declaration of LLM usage}
    \item[] Question: Does the paper describe the usage of LLMs if it is an important, original, or non-standard component of the core methods in this research? Note that if the LLM is used only for writing, editing, or formatting purposes and does \emph{not} impact the core methodology, scientific rigor, or originality of the research, declaration is not required.
    %this research? 
    \item[] Answer:  \answerNA{} % Replace by \answerYes{}, \answerNo{}, or \answerNA{}.
    \item[] Justification: LLMs were used only for language editing, typo correction, and formatting of experiment scripts.

\end{enumerate}

\end{document}